\documentclass[runningheads]{llncs}
\usepackage{cite}
\usepackage{makeidx}         
\usepackage{graphicx}        
\usepackage{multicol}   
\usepackage[bottom]{footmisc}
\usepackage{amsmath,amsfonts}
\usepackage{algorithmic}
\usepackage{array}
\usepackage{subcaption}
\usepackage{textcomp}
\usepackage{stfloats}
\usepackage{url}
\usepackage{verbatim}
\usepackage{booktabs}
\usepackage{multirow}
\usepackage{tablefootnote}
\usepackage{xcolor}
\usepackage{float}
\usepackage{lipsum}
\usepackage[english]{babel}
\usepackage{bm}
\usepackage{amssymb}
\usepackage{hyperref}
\usepackage[nameinlink]{cleveref} 

\begin{document}

\title{A Dictionary-based approach to Time Series Ordinal Classification \thanks{This work has been partially subsidised by ``Agencia Española de Investigaci\'on (España)'' (grant ref.: PID2020-115454GB-C22 / AEI / 10.13039 / 501100011033). David Guijo-Rubio's research has been subsidised by the University of Córdoba through grants to Public Universities for the requalification of the Spanish university system of the Ministry of Universities, financed by the European Union - NextGenerationEU (grant reference: UCOR01MS).}}
%
%
\author{Rafael Ayllón-Gavilán\inst{1}\orcidID{0000-0003-3024-6194} \and \\
David Guijo-Rubio\inst{1,2}\orcidID{0000-0002-8035-4057} \and \\
Pedro Antonio Gutiérrez\inst{1}\orcidID{0000-0002-2657-776X} \and \\ César Hervás-Martínez\inst{1}\orcidID{0000-0003-4564-1816}}
\authorrunning{R. Ayllón-Gavilán et al.}
%
\institute{Department of Computer Sciences, Universidad de Córdoba, 14014, Córdoba, Spain \and
School of Computing Sciences, University of East Anglia, NR4 7TQ, Norwich, United Kingdom\\
\email{\{i72aygar,dguijo,pagutierrez,chervas\}@uco.es}}
\maketitle              
%
\begin{abstract}
Time Series Classification (TSC) is an extensively researched field from which a broad range of real-world problems can be addressed obtaining excellent results. One sort of the approaches performing well are the so-called dictionary-based techniques. The Temporal Dictionary Ensemble (TDE) is the current state-of-the-art dictionary-based TSC approach. In many TSC problems we find a natural ordering in the labels associated with the time series. This characteristic is referred to as ordinality, and can be exploited to improve the methods performance. The area dealing with ordinal time series is the Time Series Ordinal Classification (TSOC) field, which is yet unexplored. In this work, we present an ordinal adaptation of the TDE algorithm, known as ordinal TDE (O-TDE). For this, a comprehensive comparison using a set of 18 TSOC problems is performed. Experiments conducted show the improvement achieved by the ordinal dictionary-based approach in comparison to four other existing nominal dictionary-based techniques.
\end{abstract}

\section{Introduction}\label{cap:sec_intro}
Machine Learning (ML) focuses on developing computer algorithms able to learn from previous experience, in such a way that they could be applied to solve real-world problems or, at least, provide support for human activities. Inside the ML paradigm, different sub-domains can be found, which emerge according to the sort of data used. Specifically, this work deals with the classification of time series. A time series is a set of values collected chronologically. This type of data can be found in a wide range of fields. For instance, the prices of a market asset over a certain period of time, or the monthly sales of a shop.

In this study, we focus on Time Series Classification (TSC), a task in which a discrete label is associated with each time series specifying some property of interest about it. The main goal is finding a model that learns the correspondence between labels and time series, so that it is capable of labelling new, unknown patterns accurately. Examples of applications can be found in medical research \cite{tsc_medicine}, psychology \cite{tsc_psicology} and industry \cite{tsc_industrial}, among others. Due to its versatility, the TSC paradigm has been greatly enhanced over the last decades. The main reason is the establishment of the UEA/UCR archive, a set of benchmark problems, that has made easier the validation of novel techniques.

TSC approaches are divided into different groups according to the methodology adopted. A first detailed taxonomy of the state of the art was presented in \cite{time_series_bakeoff}, where six main categories were distinguished: \textit{whole series}, \textit{intervals}, \textit{shapelets}, \textit{dictionary-based}, \textit{combinations} and \textit{model-based} techniques. In subsequent years, three additional groups emerged in the literature: the \textit{convolutional-based} models, introduced with the Random Convolutional Kernel Transform (ROCKET) method \cite{rocket}; \textit{deep learning-based} techniques, which mainly raised from the adaptation of residual and convolutional networks to the TSC case \cite{resnet}; and ensemble-based methods, in which the Hierarchical Vote Collective of Transformation-based Ensembles (HIVE-COTE) \cite{hc} particularly stands out due to its superiority in terms of accuracy in comparison to the rest of the state-of-the-art methodologies. Later on, an improved version of this last technique, named as HIVE-COTE 2.0 (HC2), was introduced in \cite{hc2}. The HC2 approach combines four methods from different categories: Arsenal, an ensemble of the ROCKET algorithm; Shapelet Transform Classifier (STC) \cite{stc}, a standard classifier applied to a transformation built from the distances between the phase independent subsequences, known as shapelets, and the original time series; the interval-based Diverse representation Canonical Interval Forest (DrCIF) \cite{hc2}, a random forest-based technique applied to statistical features extracted from dependent subsequences of the original time series; and the Temporal Dictionary Ensemble (TDE) \cite{tde}, an approach using bag of words representations of time series. TDE is the basis for the methodology proposed in this work.

More specifically, this work deals with the classification of ordinal time series, a special type of time series in which the associated discrete target values present a natural order relationship between them. This vaguely explored subdomain of TSC is known as Time Series Ordinal Classification (TSOC) and was firstly presented in \cite{stc_ordinal}. One example of this type of series was introduced in \cite{ethanol}, in which the task is to associate a spectrograph of 1751 observations (i.e. time series) with a label that can take four different values, \textit{E35}, \textit{E38}, \textit{E40} and \textit{E45}, ordered by the ethanol level of the sample. With this setting, during model training, misclassifying an \textit{E45} sample as \textit{E35} should be far more penalized than misclassifying it as \textit{E40}. This property is known as \textit{ordinality}, and can be exploited in a wide variety of domains including industry \cite{ordinal_for_industry_1}, image classification \cite{ordinal_for_image_detection}, atmospheric events detection \cite{ordinal_for_atmospheric_events}, finance \cite{ordinal_for_eu_sovereign_ratings}, and medicine \cite{ordinal_for_medicine}, among others. 

Finally, the goal of this work is to develop a new dictionary-based approach for the TSOC paradigm. For this, the TDE, the state-of-the-art approach in this category of TSC, is considered as the basis. For this, a TDE methodology capable of exploiting the ordinal information of the output variable is proposed. Specifically, more appropriate strategies in the ensemble member selection and in the computation of the time series symbolic representation are employed.

The remainder of this paper is organized as follows: related works are described in \Cref{cap:sec_prev}; \Cref{cap:sec_metho} describes the methodology developed, i.e. the Ordinal Temporal Dictionary Ensemble (O-TDE); \Cref{cap:sec_exper} presents the datasets and experimental settings; \Cref{cap:resul} shows the obtained results; and finally, \Cref{cap:conclusions} provides the conclusions and future research of our work.

\section{Related works}\label{cap:sec_prev}
The first dictionary-based method for time series classification was the Bag Of Patterns (BOP) presented in \cite{bop}. The BOP algorithm is divided into four phases: 1) a sliding window is applied to the time series; 2) a dimensionality reduction method called Symbolic Aggregate approXimation (SAX) \cite{sax} is used to transform each window to a symbolic representation. This representation is known as word; 3) the frequency of occurrence of each word is counted; and finally 4) histograms of words counts are computed for the time series of the training set. The prediction of new patterns is obtained through a k-Nearest Neighbours (kNN) classifier measuring the similarity between their histograms and those of the training instances.

Most of the state-of-the-art methods follow the structure of the BOP algorithm. This is the case of Bag of Symbolic Fourier approximation Symbols (BOSS) \cite{boss}. BOSS also transforms the input time series into symbolic representations (words). For this purpose, instead of SAX, it uses the Discrete Fourier Transform (DFT) \cite{dft} method. DFT avoids issues related with noisy time series, achieving a more representative transformation. 

Another distinguishing feature of BOSS is that it conforms an ensemble of BOSS approaches trained with different window sizes. Only those BOSS members achieving an over-threshold accuracy are included in the ensemble. BOSS significantly outperformed BOP. Given the performance of this approach, several BOSS-based methods were proposed in the literature. In this sense, we have the Word ExtrAction for time SEries cLassification (WEASEL) \cite{weasel} method. WEASEL applies an ANOVA test to obtain a subset of the most significant DFT coefficients for each class. From this subset it builds the bag of words for each time series. Then a chi-square test is performed to select the most significant words to compute the histograms. This feature selection methodology makes WEASEL more scalable and faster than previous proposals.

On the same line, contractable BOSS (cBOSS) \cite{sboss} performs a random selection on the parameter space making the BOSS ensemble lighter. cBOSS is significantly more scalable than BOSS but performs equally. Spatial Pyramids (SP) BOSS \cite{spboss} incorporates the SP method, widely used in computer vision problems, to the BOSS technique. SP recursively segments the input time series and computes histograms for these segments. This allows the combination of temporal and phase independent features in the symbolic transformation process, slightly improving the robustness of the algorithm.

Finally, the latest and most successful dictionary-based technique is the Temporal Dictionary Ensemble (TDE) \cite{tde}. TDE implements the same structure than BOSS, but makes use of a Gaussian process of the parameter space to do the ensemble member selection. Its superiority over competing dictionary-based methods led it to replace BOSS in the second version of the HIVE-COTE technique, HIVE-COTE2.0 (HC2) \cite{hc2}.

Focusing now on TSOC, only one type of approaches have been developed. This is the Ordinal Shapelet Transform Classifier (O-STC) \cite{stc_ordinal}. O-STC extracts phase independent features from the time series keeping those that satisfy a minimum shapelet quality (measured through a specific ordinal metric). The resulting set of shapelets are fed to an ordinal classifier such as a Proportional Odds Model (POM) \cite{pom} or an ordinal support vector machine technique \cite{svorex}. 

In this work, we focus on implementing the ordinal version of the TDE approach, given its superiority over the existing dictionary-based approaches in TSC. This technique is known as Ordinal Temporal Dictionary Ensemble (O-TDE).



\section{Ordinal Temporal Dictionary Ensemble (O-TDE)} \label{cap:sec_metho}


First of all, a time series can be categorised according to the number of dimensions $d$ as univariate ($d=1$) or multivariate ($d>1$). A univariate time series $\mathbf{x}$ of length $l$ is an ordered set of $l$ real values, $\mathbf{x}=(x_1,\ldots,x_l)$. Conversely, a multivariate time series with $d$ dimensions (or channels) and length $l$ is a collection of $d$ ordered sets, each containing $l$ real values denoted as $\mathbf{x} = \{(x_{1, 1}, \ldots, x_{1, l}), \ldots, (x_{d, 1}, \ldots x_{d, l})\}$. A time series dataset is then defined as $D=\{(\mathbf{x}_1,y_1),(\mathbf{x}_2,y_2),\ldots,(\mathbf{x}_N,y_N)\}$, where $N$ is the number of available time series, $\mathbf{x}_i$ is a time series (either univariate or multivariate), and $y_i$ is the output label associated with the respective time series. Both in this paper and in the wider TSC literature, our analyses rely on datasets comprising time series that are uniformly spaced, meaning that the observations within each time series are collected at equally-spaced time intervals. Additionally, all of the time series in the datasets are of equal length.

Focusing now on the proposal, as BOSS and TDE, O-TDE also consists of several individual techniques which, to prevent ambiguity, will be referred to as individual O-TDE. In the O-TDE algorithm, a guided parameter selection is performed to build the ensemble members. This parameter selection is guided by a Gaussian process \cite{gaussian_process} intended to predict the Mean Absolute Error (MAE) values for specific O-TDE configurations, basing its prediction on previous parameters-MAE pairs \cite{tde}. This helps to reduce the computational complexity of the ensemble construction. This process is similar to that followed in the original TDE algorithm, but considering the MAE metric instead of the accuracy. Note that MAE quantifies the error committed in the ordinal scale. Hence, it helps to boost the performance achieved for ordinal problems. 

Regarding the individual O-TDE, i.e. the method considered in the ensemble, it consists of a sequence of steps, summarised in the following lines. Firstly, a given input time series of size $l$ is processed by sliding windows of length $w$, in such a way that $w \ll l$. Then, a Discrete Fourier Transform (DFT) \cite{dft} is applied to each window, decomposing it into a set of $w$ orthogonal basis functions using sinusoidal waves. The set of waves obtained through Fourier analysis is commonly referred to as Fourier coefficients. In practice, only the first $c$ coefficients are typically retained, while the remaining coefficients, which contribute to higher frequencies, are discarded ($c \ll w$). This selection process serves two purposes: 1) since the first Fourier coefficients are related to the smoothest sections of the time series, potentially noisy parts can be eliminated. And 2) the dimensionality of the representation can be substantially reduced from $w$ coefficients to just $c$. This reduction can provide computational benefits, particularly for large or complex datasets.

At this point, from the initial time series, $c$ Fourier coefficients are kept. The $j$-th Fourier coefficient extracted from the $i$-th time series is represented by a complex number $F_{i,j} = (\text{real}_{i,j}, \text{imag}_{i,j})$. With this setting, the following matrix $A$ is built:
\begin{equation}
    A = 
    \begin{bmatrix}
        \text{real}_{1,1} & \text{imag}_{1,1} & \dotsc & \text{real}_{1,c} & \text{imag}_{1,c} \\
        \text{real}_{2,1} & \text{imag}_{2,1} & \dotsc & \text{real}_{2,c} & \text{imag}_{2,c} \\
        \vdots & \vdots & \dotsc & \vdots & \vdots \\
        \text{real}_{\text{N},1} & \text{imag}_{\text{N},1} & \dotsc & \text{real}_{\text{N},c} & \text{imag}_{\text{N},c} \\
    \end{bmatrix},
\end{equation} where $\text{N}$ is the number of time series of the training dataset. For each column of $A$, $C_m = (C_{1,m}, C_{2,m}, \dotsc, C_{N,m})$, with $m \in \{1, 2, \dotsc, 2c\}$, a set of thresholds $\bm{\beta}_m = (\beta_{m,0}, \beta_{m,1}, \dotsc, \beta_{m,T})$ is extracted through a process called Information Gain Binning (IGB) that will covered below. The $\beta_{m,0}$ and $\beta_{m,T}$ thresholds are set to $-\infty$ and $+\infty$ respectively. Note that as coefficients are represented by complex numbers (with real and imaginary parts), $m$ takes values up to $2c$. With this setting, the $C_m$ real-valued elements are discretised according to $\bm{\beta}_m$ and a finite alphabet $\bm{\Sigma} = \{\alpha_1, \alpha_2, \dotsc, \alpha_{T}\}$, where $T$ is the size of the dictionary. An element $C_{im}$ of $A$ is mapped to a symbol $\alpha_t$ of $\bm{\Sigma}$ if $\beta_{m,t-1} \leq C_{i,m} \leq \beta_{m,t}$, with $t \in \{1, 2, \dotsc, T\}$.

The resulting symbolic representation of each column is what is called a \textit{word}. The IGB process finds the optimal set $\bm{\beta}$ for each column by fitting a Decision Tree Regressor (DTR). Each $\beta_{m,i}$ corresponds to a threshold value used in a given splitting node of the tree. The impurity criterion $i$ used in the DTR is the Mean Squared Error (MSE) with an improvement score proposed in \cite{friedman_mse}:
\begin{equation}
    i = \frac{w_l \cdot w_r}{w_l + w_r} (\bar{y}_l - \bar{y}_r), 
    \label{eq:friedman-mse}
\end{equation}where $\bar{y}_l$, $\bar{y}_r$ are the left and right child nodes response means, and $w_l$, $w_r$ are the corresponding sums of the weights. The utilisation of this criterion instead of the accuracy (considered in the original TDE proposal) greatly enhances the performance in ordinal problems. This criteria is usually known in the literature as \textit{friedman-MSE}.

In base of all the above, an individual O-TDE transforms an input time series into a set of words (one word for each sliding window). Then, a histogram of words counts is built from this set. The label for a testing time series is obtained by computing the distances between its histogram and those of the training time series and returning the label of the closest one.


\section{Experimental settings}\label{cap:sec_exper}

The experiments are performed on an extended version of the TSOC archive. To avoid possible randomisation biases, $30$ runs have been performed. To measure the performance of the techniques, both nominal and ordinal metrics have been considered to get a better analysis on how the proposed ordinal methodology performs.

\subsection{Datasets considered}
With the aim of performing a robust experimentation, a set of $18$ TSOC problems from a wide variety of domains has been considered. In this section, we present these datasets and the source from which they have been collected. \Cref{tab:metadata} provides a summary of the complete set of problems. We can distinguish four different data sources: 1) The UEA/UCR TSC archive\footnote{\url{https://www.timeseriesclassification.com/dataset.php}}, where a subset of $9$ ordinal problems has been identified \cite{datasets_ordinal_UEAUCR}. 2) The Monash/UEA/UCR Time Series Extrinsic Regression (TSER) archive\footnote{\url{http://tseregression.org/}}. From this repository, we limited our selection to equal-length problems without missing values, adding two more datasets to our experiments. The originally continuous output variable of these datasets has been discretised into five equally wide bins. 3) Historical price data from $5$ of the most important companies in the stock market. We have taken this data from Yahoo Finance\footnote{\url{https://es.finance.yahoo.com/}} website, extracting weekly price data from the earliest available date to March 2023. Each time series is built with the returns over 53 weeks (the number of weeks of a year) prior to a given date $t$, and the output label corresponds to the price return in $t$ ($r_{t}$). This value is discretised according to a set of predefined symmetrical thresholds $(-\infty, -0.05, -0.02, 0.02, 0.05, \infty)$. In this way, our experimentation is extended with $5$ more problems. 4) Buoy data from the National Data Buoy Center (NDBC)\footnote{\url{https://www.ndbc.noaa.gov/}}. Two problems from this source has been considered, which are \textit{USASouthwestEnergyFlux} and \textit{USASouthwestSWH}. The first comprises a set of $468$ time series. Each time series is built on $112$ energy fluctuation measurements collected during $4$ weeks ($4$ measures per day). The objective is to estimate the level of energy fluctuation during that period of time, being $0$ the minimum level, and $3$ the highest energy level. The second problem consists on $1872$ time series of length $28$ representing sea waves height variation along a week ($4$ measures per day). The purpose is to estimate the wave height level during that period of time, ranging from $0$ (the lowest height) to $3$ (the highest height).


\newcolumntype{M}[1]{>{\arraybackslash}m{#1}}

\begin{table}[!ht]
\caption{Information about the datasets considered. OAG stands for OutlineAgeGroup.}
\begin{center}
\setlength{\tabcolsep}{2pt}
\begin{tabular}{@{}M{4.25cm}ccccc}
\toprule \toprule
Dataset name & \# Train & \# Test & \# Classes & Length & \# Dimensions \\
\midrule
AAPL                           &   1720 &   431 &        5 &      53 & 1 \\
AMZN                           &   1035 &   259 &        5 &      53 & 1 \\
AppliancesEnergy            &     95 &    42 &        5 &     144 & 24\\
AtrialFibrillation             &     15 &    15 &        3 &     640 & 2\\
Covid3Month                 &    140 &    61 &        5 &      84 & 1 \\
DistalPhalanxOAG   &    400 &   139 &        3 &      80 & 1 \\
DistalPhalanxTW                &    400 &   139 &        6 &      80 & 1 \\
EthanolConcentration           &    261 &   263 &        4 &    1751 & 3 \\
EthanolLevel                   &    504 &   500 &        4 &    1751 & 1\\
GOOG                           &    732 &   183 &        5 &      53 & 1 \\
META                           &    408 &   103 &        5 &      53 & 1 \\
MSFT                           &   1501 &   376 &        5 &      53 & 1 \\
MiddlePhalanxOAG   &    400 &   154 &        3 &      80 & 1 \\
MiddlePhalanxTW                &    399 &   154 &        6 &      80 & 1 \\
ProximalPhalanxOAG &    400 &   205 &        3 &      80 & 1 \\
ProximalPhalanxTW              &    400 &   205 &        6 &      80 & 1 \\
USASouthwestEnergyFlux      &    327 &   141 &        4 &     112 & 7 \\
USASouthwestSWH             &   1310 &   562 &        4 &      28 & 7 \\
\bottomrule \bottomrule
\end{tabular}
\end{center}

\label{tab:metadata}
\end{table}

\subsection{Experimental setup} \label{sec:experimetal_setup}
With the goal of demonstrating that ordinal approaches can outperform nominal techniques when dealing with ordinal datasets, the proposed methodology O-TDE is compared against $4$ state-of-the-art approaches in dictionary-based techniques: BOSS, cBOSS, WEASEL, and TDE. 

The performance of these approaches is measured in terms of four metrics (1 nominal and 3 ordinal). The Correct Classification Rate (CCR), also known as accuracy, is the most spread measure when dealing with nominal time series. It measures the percentage of correctly classified instances.

The first ordinal measure is the Mean Absolute Error (MAE), that quantifies the error committed in the ordinal scale:
\begin{equation}
    \label{mae}
    \text{MAE} = \frac{1}{\text{N}}\sum_{i=1}^{N}|\hat{y_i} - y_i|,
\end{equation} where $N$ represents the number of patterns, and $\hat{y}_i$ and $y_i$ are the predicted and real labels, respectively.

The second ordinal measure is the Quadratic Weighted Kappa (QWK). QWK establishes different weights depending on the different disagreement levels between real and predicted values. As MAE,
it penalises to a greater extent errors made in farther classes in the ordinal scale:
\begin{equation}
    \label{qwk}
    \text{QWK} = 1 - \frac{\sum^{\text{N}}_{i, j}\omega_{i, j}O_{i, j}}{\sum^{\text{N}}_{i, j}\omega_{i, j}E_{i, j}},
\end{equation} where $\omega$ is the penalization matrix with quadratic weights, $O$ is the confusion matrix, $E_{ij} = \frac{O_{i\bullet} O_{\bullet j}}{\text{N}}$, with $O_{i\bullet}$ and $O_{\bullet j}$ being the accumulated sum of all the elements of the $i$-th row and the $j$-th column, respectively.

The remaining ordinal metric considered is the 1-OFF accuracy (1-OFF) which is the same as the CCR but also considering as correct the predictions one category away from the actual class on the ordinal scale.

Furthermore, given that the employed methodologies have a stochastic behaviour, the experiments have been performed using $30$ different resamples. The first run is with the default data and subsequent runs are carried out with data resampled using the same train/test proportion as the original.

Finally, the code of the nominal approach is open source and is available in the \texttt{aeon} toolkit\footnote{\url{https://github.com/aeon-toolkit/aeon}}, a scikit-learn compatible implementation of the time series approaches. The ordinal version of the TDE will be included in \texttt{aeon}.


\begin{table}[!ht]
\caption{Results achieved in terms of MAE for the $5$ dictionary-based approaches considered in this work. Results are exposed as the Mean and Standard Deviation (SD) of the $30$ runs: $\text{Mean}_{\text{SD}}$.}
\setlength{\tabcolsep}{2pt}
\begin{tabular}{@{}M{3.9cm}ccccc@{}}
\toprule \toprule
Dataset &
                               \multicolumn{1}{c}{BOSS} & \multicolumn{1}{c}{cBOSS} & \multicolumn{1}{c}{WEASEL} & \multicolumn{1}{c}{TDE} & \multicolumn{1}{c}{O-TDE} \\ \midrule

AAPL & $1.383_{0.042}$ & $1.376_{0.045}$ & $\mathbf{1.292_{0.040}}$ & $1.380_{0.051}$ & $\mathit{1.364_{0.048}}$ \\
AMZN & $1.390_{0.059}$ & $1.379_{0.070}$ & $\mathbf{1.293_{0.063}}$ & $1.368_{0.068}$ & $\mathit{1.365_{0.062}}$ \\
AppliancesEnergy & $0.572_{0.010}$ & $0.571_{0.000}$ & $0.561_{0.028}$ & $\mathit{0.544_{0.044}}$ & $\mathbf{0.508_{0.058}}$ \\
AtrialFibrillation & $0.958_{0.129}$ & $\mathbf{0.802_{0.146}}$ & $0.953_{0.125}$ & $0.951_{0.178}$ & $\mathit{0.813_{0.130}}$ \\
Covid3Month & $0.755_{0.063}$ & $0.767_{0.050}$ & $0.777_{0.053}$ & $\mathbf{0.737_{0.043}}$ & $\mathit{0.747_{0.036}}$ \\
DistalPhalanxOAG & $\mathbf{0.180_{0.029}}$ & $\mathit{0.204_{0.025}}$ & $0.213_{0.025}$ & $0.207_{0.032}$ & $0.205_{0.025}$ \\
DistalPhalanxTW & $\mathit{0.386_{0.033}}$ & $0.388_{0.037}$ & $\mathbf{0.365_{0.026}}$ & $0.406_{0.030}$ & $0.404_{0.041}$ \\
EthanolConcentration & $0.725_{0.057}$ & $0.790_{0.050}$ & $\mathbf{0.513_{0.043}}$ & $0.552_{0.086}$ & $\mathit{0.539_{0.061}}$ \\
EthanolLevel & $0.561_{0.040}$ & $0.585_{0.036}$ & $\mathit{0.466_{0.061}}$ & $0.478_{0.097}$ & $\mathbf{0.425_{0.057}}$ \\
GOOG & $1.082_{0.051}$ & $1.098_{0.055}$ & $1.012_{0.055}$ & $\mathit{0.966_{0.059}}$ & $\mathbf{0.952_{0.063}}$ \\
META & $1.193_{0.082}$ & $1.185_{0.085}$ & $\mathit{1.127_{0.098}}$ & $1.150_{0.082}$ & $\mathbf{1.106_{0.072}}$ \\
MSFT & $1.101_{0.044}$ & $1.106_{0.038}$ & $\mathit{1.017_{0.042}}$ & $1.051_{0.040}$ & $\mathbf{1.009_{0.042}}$ \\
MiddlePhalanxOAG & $0.361_{0.034}$ & $0.335_{0.036}$ & $0.388_{0.039}$ & $\mathit{0.315_{0.038}}$ & $\mathbf{0.314_{0.039}}$ \\
MiddlePhalanxTW & $0.657_{0.042}$ & $0.604_{0.046}$ & $0.622_{0.046}$ & $\mathbf{0.579_{0.047}}$ & $\mathit{0.593_{0.040}}$ \\
ProximalPhalanxOAG & $0.176_{0.018}$ & $\mathbf{0.145_{0.020}}$ & $0.159_{0.018}$ & $\mathit{0.145_{0.020}}$ & $0.147_{0.020}$ \\
ProximalPhalanxTW & $0.249_{0.027}$ & $\mathit{0.216_{0.020}}$ & $0.218_{0.022}$ & $\mathbf{0.212_{0.020}}$ & $0.220_{0.017}$ \\
USASouthwestEnergy & $0.221_{0.015}$ & $0.217_{0.012}$ & $\mathbf{0.189_{0.022}}$ & $0.223_{0.025}$ & $\mathit{0.205_{0.023}}$ \\
USASouthwestSWH & $0.677_{0.048}$ & $0.391_{0.021}$ & $\mathbf{0.383_{0.013}}$ & $0.392_{0.015}$ & $\mathit{0.385_{0.014}}$ \\
\midrule
Best (second best) & $1$ ($1$) & $2$ ($2$) & $\mathit{6}$ ($3$) & $\mathit{3}$ ($4$) & $\mathbf{6}$ ($8$) \\
Rank & $4.167$ & $3.333$ & $\mathit{2.667}$ & $2.778$ & $\mathbf{2.000}$ \\
\bottomrule \bottomrule
\multicolumn{6}{l}{The best results are highlighted in bold, whereas the second-best are in italics.}
\end{tabular}\\\\

\label{tab:results_by_dataset}
\end{table}


\section{Results}\label{cap:resul}
\Cref{tab:results_by_dataset} shows the results achieved in terms of MAE. Results are shown as the 
as the mean and standard deviation of the $30$ runs carried out. As can be seen, O-TDE is the approach achieving the best results, as is the best and second best in $10$ and $4$ of the $18$ ordinal datasets, respectively. The second best approach is the nominal version of TDE, which obtained the best results for $3$ datasets (tied with WEASEL) but is the second-best in other $6$ datasets, whereas WEASEL only is the second-best in $4$.

\begin{figure}[!ht]
    \centering
    \begin{minipage}[b]{0.49\columnwidth}
        \begin{subfigure}[b]{\linewidth}
            \centering
            \includegraphics[scale = 0.46]{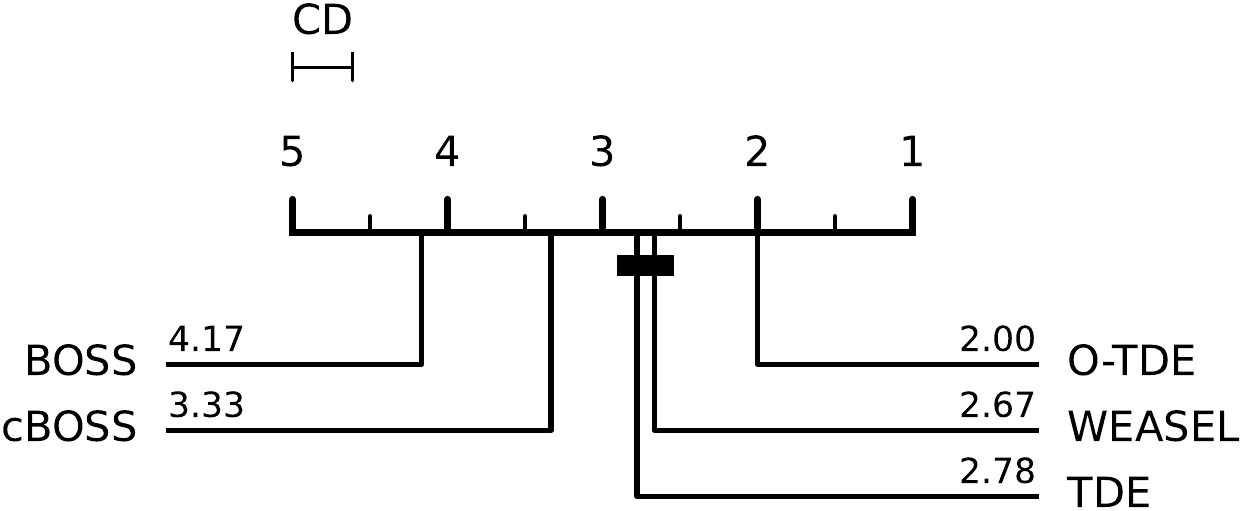}
            \caption{MAE metric.}
            \label{fig:cdd_mae}
        \end{subfigure}\vspace{8mm}

    \begin{subfigure}[b]{\linewidth}
            \centering
            \includegraphics[scale = 0.46]{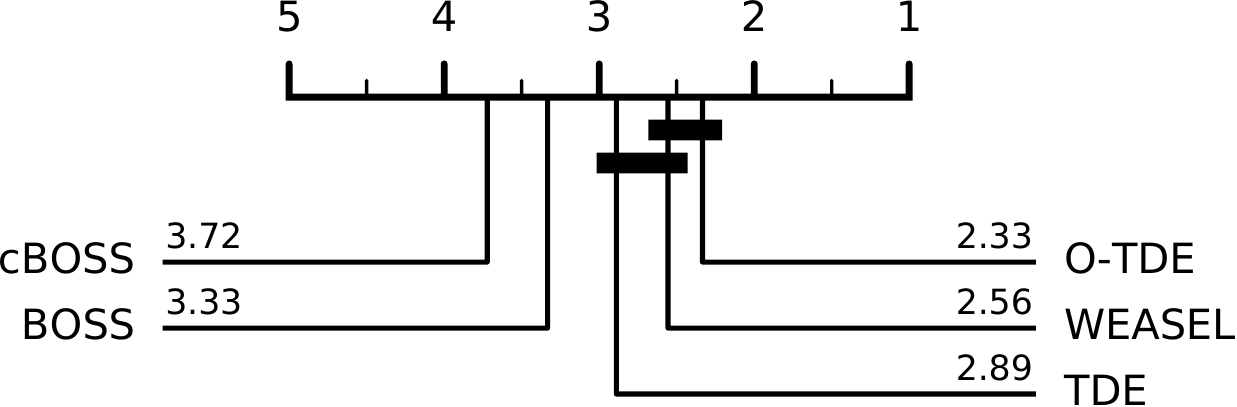}
            \caption{QWK metric.}
            \label{fig:cdd_qwk}
        \end{subfigure}
    \end{minipage}
    \begin{minipage}[b]{0.49\columnwidth}
        \centering
        \begin{subfigure}[b]{\linewidth}
            \centering
            \includegraphics[scale = 0.46]{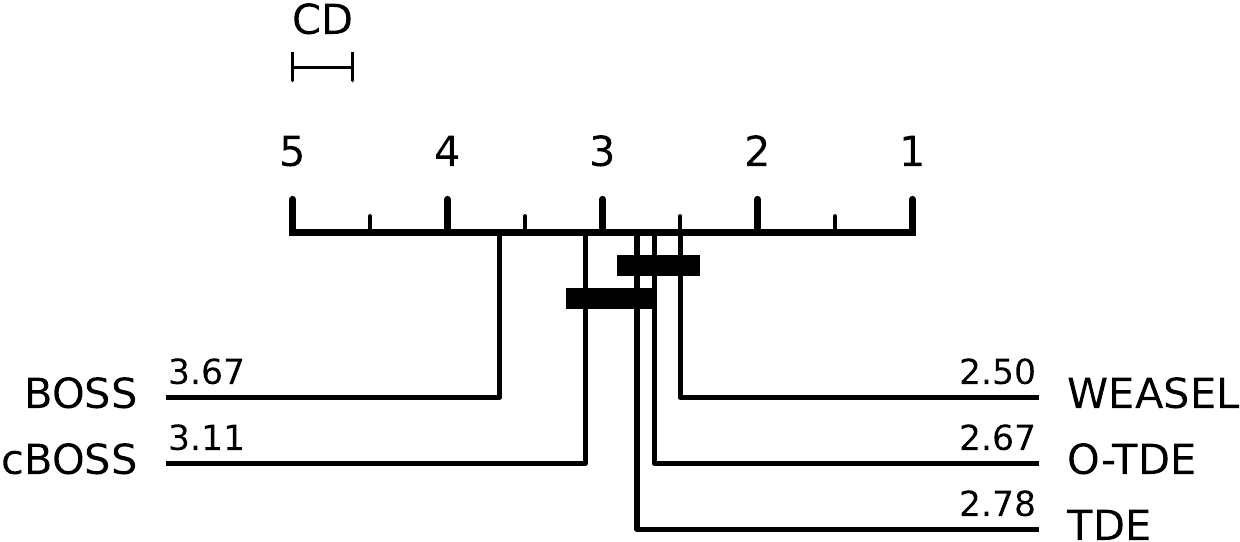}
            \caption{CCR metric.}
            \label{fig:cdd_ccr}
        \end{subfigure}\vspace{8mm}
        
        \begin{subfigure}[b]{\linewidth}
            \centering
            \includegraphics[scale = 0.46]{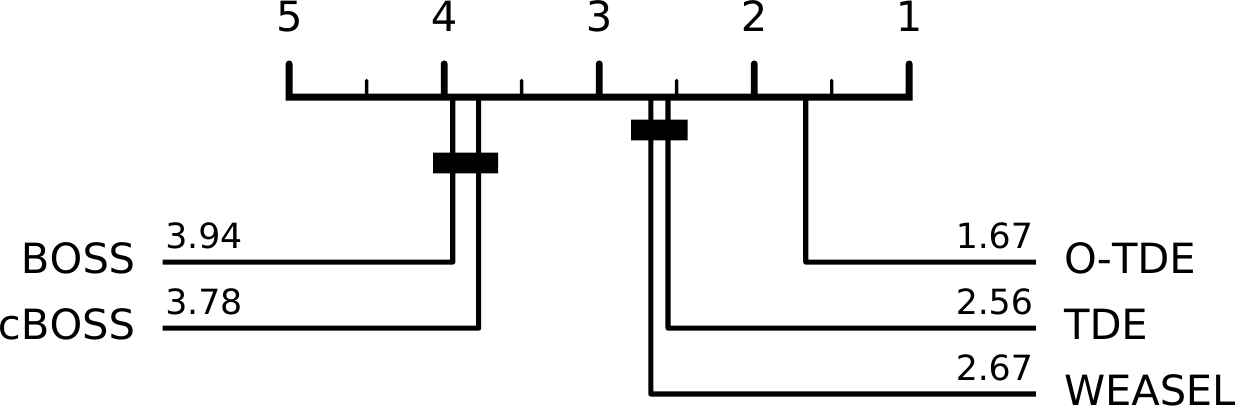}
            \caption{1-OFF metric.}
            \label{fig:cdd_off1}
        \end{subfigure}
    \end{minipage}\vspace{6mm}
    
    \caption{CDDs in terms of MAE (a), QWK (b), CCR (c) and 1-OFF (d). The significance value $\alpha$ is set to $0.1$. The critical difference (CD) value is computed pairwise and is equal to $0.456$.}
    \label{cdd}
\end{figure}

Furthermore, to compare the results obtained for multiple classifiers over multiple datasets, Critical Difference Diagrams (CDDs) are used \cite{critical_difference_diagram}. The post-hoc Nemenyi test is replaced by a comparison of all classifiers using pairwise Wilcoxon signed-rank tests. Finally, cliques are formed using the Holm correction \cite{holm_cdd}. \Cref{cdd} shows the CDDs for the four measures detailed in \Cref{sec:experimetal_setup}.

From these results, it can be said that a solid superiority of the O-TDE method is observed against the nominal methodologies. O-TDE outperforms all the nominal techniques not only in terms of ordinal performance measures (MAE and QWK) but also in terms of CCR, a nominal measure. Even though improving the results in CCR is not the final goal of the ordinal approaches, this superiority demonstrates the potential of the ordinal techniques over nominal ones. Finally, indicate that this difference becomes statistically significant for the MAE and 1-OFF metrics, indicating an excellent performance of the O-TDE proposed approach.

\section{Conclusion and future scope}\label{cap:conclusions}
Time Series Ordinal Classification is still an unexplored paradigm in the time series literature, being a subset of the popular nominal Time Series Classification (TSC) task. However, it has a wealth of real-world applications in a wide range of fields such as finances, medicine or energy, among others. In this work, it has been shown that when this sort of problems are approximated through ordinal methods, such as the presented Ordinal Temporal Dictionary Ensemble (O-TDE), a significant boost in performance is obtained. This superiority is mainly achieved by penalising more severely those predictions that fall far away from the real class in the ordinal scale. 

From the original set of $7$ datasets previously identified, this work provides another $11$ datasets, taking the ordinal archive to 18 ordinal datasets, including $13$ univariate and $5$ multivariate, making the obtained results more robust. The performance of the $5$ approaches has been measured in terms of accuracy, the most used one in nominal TSC, and three ordinal metrics, Mean Average Error (MAE), Quadratic Weighted Kappa (QWK) and 1-OFF accuracy (1-OFF). These three measures help to properly quantify the capacity of the approaches to model the ordinal scale. Consequently, the biggest differences in performance between nominal and ordinal methodologies are obtained in terms of these last three metrics, being the difference in terms of MAE and 1-OFF statistically significant. 

For future works, the TSOC archive is sought to be expanded. In addition, multiple well-known TSC methods such as kernel-based, ensemble-based or interval-based techniques will be explored for the ordinal paradigm.
\bibliographystyle{splncs04}
\bibliography{bibliography}

\begin{thebibliography}{10}
\providecommand{\url}[1]{\texttt{#1}}
\providecommand{\urlprefix}{URL }
\providecommand{\doi}[1]{https://doi.org/#1}

\bibitem{time_series_bakeoff}
Bagnall, A., Lines, J., Bostrom, A., Large, J., Keogh, E.: The great time
  series classification bake off: a review and experimental evaluation of
  recent algorithmic advances. DATA MINING AND KNOWLEDGE DISCOVERY
  \textbf{31}(3),  606--660 (MAY 2017)

\bibitem{holm_cdd}
Benavoli, A., Corani, G., Mangili, F.: Should we really use post-hoc tests
  based on mean-ranks? The Journal of Machine Learning Research
  \textbf{17}(1),  152--161 (2016)

\bibitem{tsc_medicine}
Buza, K., Koller, J., Marussy, K.: Process: projection-based classification of
  electroencephalograph signals. In: Artificial Intelligence and Soft
  Computing: 14th International Conference, ICAISC 2015, Zakopane, Poland, June
  14-18, 2015, Proceedings, Part II 14. pp. 91--100. Springer (2015)

\bibitem{svorex}
Chu, W., Keerthi, S.S.: New approaches to support vector ordinal regression.
  In: Proceedings of the 22nd international conference on Machine learning. pp.
  145--152 (2005)

\bibitem{rocket}
Dempster, A., Petitjean, F., Webb, G.I.: Rocket: exceptionally fast and
  accurate time series classification using random convolutional kernels. Data
  Mining and Knowledge Discovery  \textbf{34},  1454--1495 (2020)

\bibitem{critical_difference_diagram}
Dem{\v{s}}ar, J.: Statistical comparisons of classifiers over multiple data
  sets. The Journal of Machine learning research  \textbf{7},  1--30 (2006)

\bibitem{ordinal_for_eu_sovereign_ratings}
Fernandez-Navarro, F., Campoy-Munoz, P., de~la Paz-Marin, M., Hervas-Martinez,
  C., Yao, X.: Addressing the eu sovereign ratings using an ordinal regression
  approach. IEEE Transactions on Cybernetics  \textbf{43}(6),  2228--2240 (DEC
  2013)

\bibitem{friedman_mse}
Friedman, J.H.: Greedy function approximation: A gradient boosting machine. The
  Annals of Statistics  \textbf{29}(5),  1189--1232 (2001)

\bibitem{ordinal_for_atmospheric_events}
Guijo-Rubio, D., Casanova-Mateo, C., Sanz-Justo, J., Gutierrez, P.,
  Cornejo-Bueno, S., Herv{\'a}s, C., Salcedo-Sanz, S.: Ordinal regression
  algorithms for the analysis of convective situations over madrid-barajas
  airport. Atmospheric Research  \textbf{236},  104798 (2020)

\bibitem{datasets_ordinal_UEAUCR}
Guijo-Rubio, D., Guti{\'e}rrez, P.A., Bagnall, A., Herv{\'a}s-Mart{\'\i}nez,
  C.: Ordinal versus nominal time series classification. In: Advanced Analytics
  and Learning on Temporal Data: 5th ECML PKDD Workshop, AALTD 2020, Ghent,
  Belgium, September 18, 2020, Revised Selected Papers. pp. 19--29. Springer
  (2020)

\bibitem{stc_ordinal}
Guijo-Rubio, D., Gutiérrez, P.A., Bagnall, A., Hervás-Martínez, C.: Time
  series ordinal classification via shapelets. In: 2020 International Joint
  Conference on Neural Networks (IJCNN). pp.~1--8 (2020)

\bibitem{dft}
Harris, F.J.: On the use of windows for harmonic analysis with the discrete
  fourier transform. Proceedings of the IEEE  \textbf{66}(1),  51--83 (1978)

\bibitem{stc}
Hills, J., Lines, J., Baranauskas, E., Mapp, J., Bagnall, A.: Classification of
  time series by shapelet transformation. Data mining and knowledge discovery
  \textbf{28}(4),  851--881 (2014)

\bibitem{tsc_psicology}
Kurbalija, V., von Bernstorff, C., Burkhard, H.D., Nachtwei, J., Ivanovi{\'c},
  M., Fodor, L.: Time-series mining in a psychological domain. In: proceedings
  of the Fifth Balkan Conference in Informatics. pp. 58--63 (2012)

\bibitem{spboss}
Large, J., Bagnall, A., Malinowski, S., Tavenard, R.: On time series
  classification with dictionary-based classifiers. Intelligent Data Analysis
  \textbf{23}(5),  1073--1089 (2019)

\bibitem{ethanol}
Large, J., Kemsley, E.K., Wellner, N., Goodall, I., Bagnall, A.: Detecting
  forged alcohol non-invasively through vibrational spectroscopy and machine
  learning. In: Advances in Knowledge Discovery and Data Mining: 22nd
  Pacific-Asia Conference, PAKDD 2018, Melbourne, VIC, Australia, June 3-6,
  2018, Proceedings, Part I 22. pp. 298--309. Springer (2018)

\bibitem{sax}
Lin, J., Keogh, E., Wei, L., Lonardi, S.: Experiencing sax: a novel symbolic
  representation of time series. Data Mining and knowledge discovery
  \textbf{15},  107--144 (2007)

\bibitem{bop}
Lin, J., Khade, R., Li, Y.: Rotation-invariant similarity in time series using
  bag-of-patterns representation. Journal of Intelligent Information Systems
  \textbf{39},  287--315 (2012)

\bibitem{hc}
Lines, J., Taylor, S., Bagnall, A.: Time series classification with hive-cote:
  The hierarchical vote collective of transformation-based ensembles. ACM
  Transactions on Knowledge Discovery from Data  \textbf{12}(5) (2018)

\bibitem{ordinal_for_image_detection}
Liu, Y., Wang, Y., Kong, A.W.K.: Pixel-wise ordinal classification for salient
  object grading. Image and Vision Computing  \textbf{106} (FEB 2021)

\bibitem{tsc_industrial}
Malhotra, P., Vig, L., Shroff, G., Agarwal, P., et~al.: Long short term memory
  networks for anomaly detection in time series. In: ESANN. vol.~2015, p.~89
  (2015)

\bibitem{pom}
McCullagh, P.: Regression models for ordinal data. Journal of the Royal
  Statistical Society: Series B (Methodological)  \textbf{42}(2),  109--127
  (1980)

\bibitem{tde}
Middlehurst, M., Large, J., Cawley, G., Bagnall, A.: The temporal dictionary
  ensemble (tde) classifier for time series classification. In: Machine
  Learning and Knowledge Discovery in Databases: European Conference, ECML PKDD
  2020, Ghent, Belgium, September 14--18, 2020, Proceedings, Part I. pp.
  660--676. Springer (2021)

\bibitem{hc2}
Middlehurst, M., Large, J., Flynn, M., Lines, J., Bostrom, A., Bagnall, A.:
  Hive-cote 2.0: a new meta ensemble for time series classification. MACHINE
  LEARNING  \textbf{110}(11-12),  3211--3243 (DEC 2021)

\bibitem{sboss}
Middlehurst, M., Vickers, W., Bagnall, A.: Scalable dictionary classifiers for
  time series classification. In: Yin, H., Camacho, D., Tino, P.,
  Tall{\'o}n-Ballesteros, A.J., Menezes, R., Allmendinger, R. (eds.)
  Intelligent Data Engineering and Automated Learning -- IDEAL 2019. pp.
  11--19. Springer International Publishing, Cham (2019)

\bibitem{boss}
Sch{\"a}fer, P.: The boss is concerned with time series classification in the
  presence of noise. Data Mining and Knowledge Discovery  \textbf{29},
  1505--1530 (2015)

\bibitem{weasel}
Sch{\"a}fer, P., Leser, U.: Fast and accurate time series classification with
  weasel. In: Proceedings of the 2017 ACM on Conference on Information and
  Knowledge Management. pp. 637--646 (2017)

\bibitem{gaussian_process}
Schulz, E., Speekenbrink, M., Krause, A.: A tutorial on gaussian process
  regression: Modelling, exploring, and exploiting functions. Journal of
  Mathematical Psychology  \textbf{85},  1--16 (2018)

\bibitem{ordinal_for_industry_1}
Vargas, V.M., Guti{\'e}rrez, P.A., Rosati, R., Romeo, L., Frontoni, E.,
  Herv{\'a}s-Mart{\'\i}nez, C.: Deep learning based hierarchical classifier for
  weapon stock aesthetic quality control assessment. Computers in Industry
  \textbf{144},  103786 (2023)

\bibitem{resnet}
Wang, Z., Yan, W., Oates, T.: Time series classification from scratch with deep
  neural networks: A strong baseline. In: 2017 International joint conference
  on neural networks (IJCNN). pp. 1578--1585. IEEE (2017)

\bibitem{ordinal_for_medicine}
Zhou, Z., Huang, B., Zhang, R., Yin, M., Liu, C., Liu, Y., Yi, Z., Wu, X.:
  Methods to recognize depth of hard inclusions in soft tissue using ordinal
  classification for robotic palpation. IEEE Transactions on Instrumentation
  and Measurement  \textbf{71},  1--12 (2022)

\end{thebibliography}

\end{document}